# Latent Composite Likelihood Learning for the Structured Canonical Correlation Model

**Ricardo Silva**
Department of Statistical Science and CSML
University College London
RICARDO@STATS.UCL.AC.UK

## Abstract

Latent variable models are used to estimate variables of interest – quantities which are observable only up to some measurement error. In many studies, such variables are known but not precisely quantifiable (such as "job satisfaction" in social sciences and marketing, "analytical ability" in educational testing, or "inflation" in economics). This leads to the development of measurement instruments to record noisy indirect evidence for such unobserved variables such as surveys, tests and price indexes. In such problems, there are postulated latent variables and a given measurement model. At the same time, other unantecipated latent variables can add further unmeasured confounding to the observed variables. The problem is how to deal with unantecipated latents variables. In this paper, we provide a method loosely inspired by canonical correlation that makes use of background information concerning the "known" latent variables. Given a partially specified structure, it provides a structure learning approach to detect "unknown unknowns," the confounding effect of potentially infinitely many other latent variables. This is done without explicitly modeling such extra latent factors. Because of the special structure of the problem, we are able to exploit a new variation of composite likelihood fitting to efficiently learn this structure. Validation is provided with experiments in synthetic data and the analysis of a large survey done with a sample of over 100,000 staff members of the National Health Service of the United Kingdom.

## 1 CONTRIBUTION

We present a method for learning the structure of a latent variable model, where latent variables are divided into two categories: i. latent variables which we would like to estimate, as in any smoothing task (e.g., to generate latent representations of data points for visualization, clustering, and ranking, among other tasks); ii. all other latent variables, which we are not interested in estimating but which can add further confounding among observed variables and as such cannot be ignored. They are nuisance variables.

This setup is motivated by many practical problems in the applied sciences where target latent variables are chosen upfront, with observed variables designed to measure the unobservable variables of interest. Consider the simple illustrative example of Figure 1.

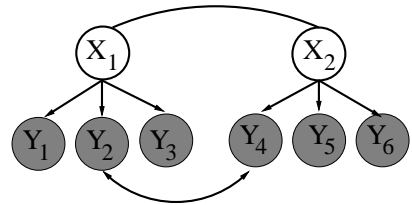

Figure 1: A latent variable model with a mixed graphical representation (that is, more than one type of edge).

Here, each $X_i$ is a latent variable and each $Y_i$ an observed variable. For instance, $X_1$ could represent a latent variable quantifying the level of job satisfaction, postulated to be measured by answers to questions such as $Y_1 \equiv$ "I can choose my own method of working" and $Y_2 \equiv$ "I have strong support from my manager". $X_2$ could represent a latent level of job responsibility, measured by questions such as $Y_4 \equiv$ "I know what my responsibilities are" and $Y_5 \equiv$ "I am regularly consulted by my team".

Other nuisance latent factors might correlate, say, the employee support from her manager and the level of interaction with her team. That is, associations not accounted by $X_1$ and $X_2$. This is represented graphically in Figure 1 by a bi-directed edge between two observed items, a notation widely used in the structural equation model literature (Bollen, 1989) and futher formalized by Richardson and Spirtes (2002). In a linear Gaussian model, for in-stance, this could be parameterized by the measurement model equation $Y_i = \mathbf{X}^\mathsf{T}\beta + \epsilon_i$, where the covariance of the error terms for $Y_2$ and $Y_4$ is allowed to be non-zero. More details will be given in the sequel.

Conditioned on the background knowledge of a model where we specify the "known unknowns" – the latent features we would like to estimate by a postulated relationship to observations – we would like to identify the remaining dependence structure given by all remaining factors we did not specify in advance – the "unknown unknowns." This provides a meeting point between the common practice of designing measurement models for the practical goal of quantifying specific latent factors (Bollen, 1989), and purely data-driven approaches for discovering latent structure (Elidan et al., 2000; Silva et al., 2006).

The structure of the paper is as follows. A formal problem definition is given in Section 2, where we specifiy precisely the goals, classes of models and assumptions embedded in our procedure. Algorithms for structure learning are described in detail in Section 3, based on variations of the composite likelihood method integrated within a expectation-maximization framework. Further context and related work is discussed in Section 4. Experiments with synthetic and real data are discussed in Section 5, followed by a Conclusion in Section 6.

## 2 PROBLEM SPECIFICATION AND MODEL SPACE

In this section, we first define the space of graphical models which is assumed to contain the data generating process of interest. We then present further details on the parametric assumptions and a formulation based on *cumulative distribution networks* (Huang and Frey, 2008; Silva et al., 2011).

### 2.1 Structural Conditions

In our setup, we assume the following:

1. that the data is generated according to $\mathcal{G}$, an underlying directed acyclic graph (DAG) (Koller and Friedman, 2009) composed of latent and observed variables. In $\mathcal{G}$, observed variables are not parents of any other variable, as in typical models of factor analysis (Bartholomew and Knott, 1999);
2. that an expert provides a partition of the set of observed variables $\mathbf{Y}$, such that each set $\mathcal{S}_i \subset \mathbf{Y}$ in this partition corresponds to the observed children of a latent variable $X_i$. For example, in Figure 1, the partition is given by sets $\mathcal{S}_1 \equiv \{Y_1, Y_2, Y_3\}$ and $\mathcal{S}_2 \equiv \{Y_4, Y_5, Y_6\}$;
3. let $\mathbf{X}_\mathcal{S}$ be this set of latent variables associated with the partition (e.g., $\{X_1, X_2\}$ in Figure 1). We also allow for a (possibly infinite) set of latent variables $\mathbf{X}_\infty$ that is disjoint of $\mathbf{X}_\mathcal{S}$;
4. elements of $\mathbf{X}_\mathcal{S}$ are arbitrarily connected to each other in $\mathcal{G}$, and elements of $\mathbf{X}_\infty$ are also arbitrarily connected to each other. However, any $X_i \in \mathbf{X}_\mathcal{S}$ is marginally independent of any $X_j \in \mathbf{X}_\infty$ in $\mathcal{G}$.

### 2.2 Problem Specification

We want to account for any association among elements of $\mathbf{Y}$ that is due to $\mathbf{X}_\infty$, so that functionals of the conditional distribution $\mathcal{P}(\mathbf{X}_\mathcal{S} \mid \mathbf{Y})$ can be correctly estimated from data.

Instead of modeling the number of elements in $\mathbf{X}_\infty$ and how they relate to each other as in a nonparametric latent variable model formulation (e.g. Wood et al., 2006), we will treat the existence of $\mathbf{X}_\infty$ as a black-box. We model directly the dependencies that arise from its existence within a *mixed graph* formulation (Richardson and Spirtes, 2002). A mixed graph is a graph with more than one type of edge.

Let $\mathcal{G}_m$ be a graph with vertices $\mathbf{Y} \cup \mathbf{X}_\mathcal{S}$, and directed edges from each $X_i$ to each element of $\mathcal{S}_i$. For simplicity, the structure among elements of $\mathbf{X}_\mathcal{S}$ is assumed to be a fully connected network of undirected edges. If two observed variables $Y_i$ and $Y_j$ have a common ancestor in $\mathbf{X}_\infty$ in $\mathcal{G}$, add a bi-directed edge $Y_i \leftrightarrow Y_j$ to $\mathcal{G}_m$. We say that $\mathcal{G}_m$ is the mixed graph induced by $\mathcal{G}$ and $\mathbf{X}_\mathcal{S}$. The conditional independence constraints entailed by $\mathcal{G}_m$ all hold in $\mathcal{G}$ (Richardson and Spirtes, 2002). Hence, identifiability issues aside, fitting a model based on a graph different from $\mathcal{G}_m$ should in general result in a misspecified distribution and give a inadequate model for $\mathcal{P}(\mathbf{X}_\mathcal{S} \mid \mathbf{Y})$.

Our goal can then be summarized as: *learn the structure of $\mathcal{G}_m$ by finding the correct bi-directed substructure.* In the example of Figure 1, this means returning the information that $Y_2 \leftrightarrow Y_4$ is the only bi-directed edge in $\mathcal{G}_m$.

The *practical* assumption here is that the bi-directed component of $\mathcal{G}_m$ is sparse. Sparsity in a bi-directed component corresponds to a marginal independence constraint among elements of $\mathbf{X}_\infty$: the lack of an edge $Y_i \leftrightarrow Y_j$ corresponds to the parents of $Y_i$ in $\mathbf{X}_\infty$ being marginally independent of the parents of $Y_j$ in $\mathbf{X}_\infty$. The extent to which this assumption is valid will depend on how well the dependence between elements of $\mathbf{Y}$ is captured by $\mathbf{X}_\mathcal{S}$ and it is verified empirically in Section 5. This complements Wood et al. (2006), which makes fewer assumptions but has to deal with a much harder problem.

### 2.3 Comments

The setup where observed variables are partitioned into sets corresponding to particular semantic groups is a key idea behind *canonical correlation analysis* (CCA). The standard CCA corresponds to a partition into two sets. For rank-one CCA, this corresponds to a graphical model with a single latent variable being a parent of all observations, variables in each partition connected to each other by bi-directed edges. There are no bi-directed connections across variables in different sets of the partition, but CCA in general allows for several independent latent variables being common parents of all observations. A general latent variable model interpretation of canonical correlation analysis was originally introduced in a series of reports by Wegelin et al. (2001, 2002), and by Bach and Jordan (2005). In contrast, we are assuming that a single latent variable is (explicitly) a parent of each given group of variables, that such latent variables can be mutually dependent, but that extra factors are allowed. Common between our model space and CCA is the idea of having a partition of the observed variables. As such, we refer to the problem of searching for a structure in the space of graphs with a known partition as the *structured canonical correlation analysis* problem. We once again emphasize that a major motivation for this type of analysis is a domain-dependent assumption about which latent variables are being measured, and that such variables have a prior interpretation.

### 2.4 Parametric Assumptions

For a fixed level of $\mathbf{X}_\mathcal{S}$, observed variables $\mathbf{Y}$ should have marginal independence constraints as implied by the corresponding bi-directed structure. In a Gaussian parameterization, for instance, this would mean that the covariance of $Y_i$ and $Y_j$ given $\mathbf{X}_\mathcal{S}$ is constrained to be zero if $Y_i$ is not adjacent to $Y_j$, and free otherwise (Richardson and Spirtes, 2002).

For this paper, we will model binary data. Latent variables $\mathbf{X}_\mathcal{S}$ are assumed to follow a zero mean Gausssian with an unknown covariance matrix $\Sigma$. For the rest of this section, we discuss models for the conditional distribution of observed variables given $\mathbf{X}_\mathcal{S}$. We will assume a linear model for each univariate dependence $\mathcal{P}(Y_i = 0 \mid X_j = x_j) \equiv \Phi(0; \beta_{i1}x_j + \beta_{i0}, 1)$ for $Y_i \in \mathcal{S}_j$. Function $\Phi(0; \mu, \sigma^2)$ is the probability of a Gaussian of mean $\mu$ and variance $\sigma^2$ being negative. Coefficients $\{\beta_{i1}, \beta_{i0}\}$ are unknown.

Given these univariate conditionals, a conditional joint is needed. Models of marginal independence for binary data can be obtained from a Gaussian parameterization, as discussed by Silva and Ghahramani (2009). However, in our context we will be also interested in performing Bayesian model selection. Although priors for sparse covariance matrices exist, performing model selection in a large space is computationally problematic: we are particularly motivated by the modeling of surveys with a large sample size. More details on that are discussed in Section 4.

#### 2.4.1 Brief review of CDNs

Instead, we will adopt the cumulative distribution network framework (CDN) of Huang and Frey (2008, 2011). Given a symmetric graph, a CDN model defines the cumulative distribution function $F(\mathbf{y})$ of a multivariate distribution by a product of $f$ factors. Each factor $F_i(\cdot)$ has as arguments the variables in a clique $\mathbf{Y}_{f_i}$ of the graph:

$$F(\mathbf{y}) \equiv \prod_{i=1}^{f} F_i(\mathbf{y}_{f_i})$$

As shown by Silva et al. (2011), this can be extended to accommodate conditional distributions and integrated with a copula modeling framework (Nelsen, 2007) [1]. For a fixed instantiation $\mathbf{x}$ of the parents $\mathbf{X}$ of a set of random variables $\mathbf{Y}$, its conditional CDF $\mathcal{P}(\mathbf{Y} \leq \mathbf{y} \mid \mathbf{X} = \mathbf{x}) \equiv F(\mathbf{y} \mid \mathbf{x})$ can be parameterized as

$$F(\mathbf{y} \mid \mathbf{x}) \equiv \prod_{i=1}^{f} C_i(u_\mathbf{x}(y_{f_i;1})^{e_{f_i;1}}, \ldots, u_\mathbf{x}(y_{f_i;n(i)})^{e_{f_i;n(i)}}) \quad (1)$$

where $C_i(\cdot)$ is a copula function, $y_{f_i;j}$ denotes the instantiation of the $j$-th variable in clique $f_i$, exponent $e_{f_i;j}$ is a non-negative parameter, and $n(i)$ the number of elements in clique $f_i$. Moreover,

$$u_\mathbf{x}(y_i) \equiv \mathcal{P}(Y_i \leq y_i \mid \mathbf{x}) \quad (2)$$

Also, for each $Y_i$, the sum of its respective exponents $e_\star$, across all factors containing $Y_i$, is equal to 1. For simplicity, we fix each exponent associated with $Y_i$ to be $1/h(i)$, with $h(i)$ being the number of factors containing $Y_i$.

A powerful property of CDNs is its simple marginalization procedure: calculation of the marginal of a CDF by marginalizing a subset $\mathbf{Y}_{marg}$ consists of evaluating the CDF where $\mathbf{Y}_{marg} = \infty$. For instance, $\mathcal{P}(Y_1 \leq y_1, Y_2 \leq y_2) = \mathcal{P}(Y_1 \leq y_1, Y_2 \leq y_2, Y_3 \leq \infty, \ldots, Y_N \leq \infty)$.

In a graph such as $Y_1 \leftrightarrow Y_2 \leftrightarrow Y_3$, the CDF is given by a CDN model $F_1(y_1, y_2)F_2(y_2, y_3)$ and the corresponding marginal for $Y_1$ and $Y_3$ is given by $F_1(y_1, \infty)F_2(\infty, y_3) \equiv g(y_1)h(y_3)$, corroborating the fact that $Y_1$ and $Y_3$ should be marginally independent. In the case of a model for $F(y_1, y_2, y_3 \mid \mathbf{x})$, according to (1), we have that $F(y_1, y_2, y_3 \mid \mathbf{x})$ is given by

$$C_1(u_\mathbf{x}(y_1), u_\mathbf{x}(y_2)^{1/2})C_2(u_\mathbf{x}(y_2)^{1/2}, u_\mathbf{x}(y_3))$$

---

[1] A full description of copula models is beyond the scope of this paper. Nelsen (2007) provides further details, and Huang and Frey (2011), Silva et al. (2011) provide examples within the context of CDNs. For the purposes of this paper, it suffices to understand that a copula is nothing but a $d$-dimensional CDF with (continuous) uniform marginals in $[0, 1]^d$. Its motivation can then be understood: it allows for the construction of a joint distribution where marginals can be parameterized separately. By defining each univariate marginal $\mathcal{P}(y_i \mid \mathbf{x})$ separately, the transformation $u_\mathbf{x}(y_i) = \mathcal{P}(Y_i \leq y_i \mid \mathbf{x})$ gives an uniform $[0, 1]$ random variable. These transformed variables can then be plugged into a joint distribution with uniform marginals, which becomes a distribution with arbitrary marginals in the original space. This, among other uses, allows for the use of plug-in estimates for marginals to be combined with other estimators for joints. In this case, no further bias will be introduced into the marginal models even if the joint is misspecified or if the copula estimator is biased.

Recall that, if $C(u, v)$ is a copula function, then $C(u, 1) = u$ (since copulas are CDFs with $U(0, 1)$ marginals). One can then verify that $F(y_1, y_3 \mid \mathbf{x}) = \mathcal{P}(Y_1 \leq y_1 \mid \mathbf{x})\mathcal{P}(Y_3 \leq y_3 \mid \mathbf{x})$ and $F(y_2 \mid \mathbf{x}) = \mathcal{P}(Y_2 \leq y_2 \mid \mathbf{x})$ as desired.

Another advantage of the CDN formulation is that parameters across factors are functionally independent. This is in contrast with, for instance, the sparse covariance models of Richardson and Spirtes (2002) and Silva and Ghahramani (2009), where a positive definite constraint ties all parameters. Having no constraints across different factors will be a fundamental property to be exploited in our learning procedure. We call this property *parameter modularity*.

The difficulty with the CDN formulation is that in order to calculate the likelihood function, as required for any likelihood-based learning procedure, one has to convert the conditional CDFs into probability mass functions (PMFs) (Huang and Frey, 2011). This can potentially take an exponential amount of time on the number of variables in the graph. However, the marginalization property and the parameter modularity property of the CDNs leads to an attractive way of performing efficient learning, as we will see in the next section.

## 3 A LATENT COMPOSITE LIKELIHOOD APPROACH

Let $\mathcal{D} = \{\mathbf{Y}^{(1)}, \ldots, \mathbf{Y}^{(N)}\}$ be our data, a set of binary measurements of dimension $p$. For a given graphical structure $\mathcal{G}_m$ and fixed coefficient parameters $\{\beta_{i1}, \beta_{i0}\}$ and latent covariance matrix $\Sigma$, the marginal likelihood of $\{\mathcal{G}_m, \{\beta_{i1}, \beta_{i0}\}, \Sigma\}$ is

$$\mathcal{P}(\mathcal{D} \mid \mathcal{G}_m, \{\beta_{i1}, \beta_{i0}\}, \Sigma) =$$

$$\int \mathcal{P}(\mathcal{D}, \mathbf{X}^{1:N}, \theta \mid \mathcal{G}_m, \beta, \Sigma) \, d\mathbf{X}^{1:N} \, d\theta \quad (3)$$

where $\mathbf{X}^{1:N}$ is a shorthand notation for $\{\mathbf{X}^{(1)}, \ldots, \mathbf{X}^{(N)}\}$, $\beta$ is a shorthand notation for $\{\beta_{1,1}, \beta_{1,0}, \ldots, \beta_{p,1}, \beta_{p,0}\}$, and parameter set $\theta$ describes parameters associated with copula functions.

One approach for model selection is to assign a prior $\pi(\mathcal{G}_m)$ over possible graph structures and then choose $\mathcal{G}_m$ that maximizes $\mathcal{P}(\mathcal{D} \mid \mathcal{G}_m, \{\beta_{i1}, \beta_{i0}\}, \Sigma)\pi(\mathcal{G}_m)$. Since $\{\{\beta_{i1}, \beta_{i0}\}, \Sigma\}$ are also unknown, we could marginalize then away. However, such parameters do not affect the complexity of the model, and with the goal of having an efficient computational procedure, we will treat these parameters also as nuisance parameters and maximize with respect to them along with $\mathcal{G}_m$. In this paper, we will assume that individual pairwise factors are associated with each bi-directed edge in $\mathcal{G}_m$. Our prior for $\theta$ is independent[2] of $\mathcal{G}_m$ and factorizes as

$$\pi(\theta \mid \mathcal{G}_m) = \pi(\theta) = \prod_{1 \leq i < j \leq p} \pi(\theta_{ij}) \quad (4)$$

Maximizing any expression that depends on (3) poses a formidable computational problem. Moreover, it depends on conditional probability expressions $\mathcal{P}(\mathbf{Y} \mid \mathbf{x})$. Such conditional mass functions need to be obtained from the canonical CDF $F(\cdot)$ to PMF $\mathcal{P}(\cdot)$ transformation (Joe, 1997), which for binary data boils down to:

$$\mathcal{P}(\mathbf{Y} = \mathbf{y}) = \sum_{z_1=0}^{1} \cdots \sum_{z_p=0}^{1} (-1)^{\sum_{i=1}^{p} z_i} F(\mathbf{y} - \mathbf{z}) \quad (5)$$

This is of course exponential in $p$, but if the corresponding bi-directed component is a symmetric graph of low tree-width, the expression can be calculated efficiently by dynamic programming and used by any learning method that requires the likelihood function. The original derivation by Huang et al. (2010) is quite complex and provides insights on how approximate methods should behave. However, in the sequel we will make use of exact methods only, and as such we provide in the Supplementary Material a straightforward reduction of (5) to a standard inference problem in factor graphs – making the method easier to implement.

Integrating away $\theta$ and $\mathbf{X}$ is harder in general and large tree-width graphs are still a possibility. As such we avoid methods that attempt to maximize (3). The core procedure is based on structural composite likelihood learning. It is described in Section 3.1, and refined in Section 3.2. Further implementation details are given in Section 3.3. An alternative to these methods is to emulate a constraint-based approach (Spirtes et al., 2000) for structure discovery, providing a non-iterative procedure to identify which bi-directed edges are needed. This is done in Section 3.4. A brief discussion on model identification is provided in the Supplementary Material.

### 3.1 Basic Structural Composite Likelihood

A composite likelihood function (Varin et al., 2011) for a parameter of interest $\theta$ is defined as

$$CL(\theta; \mathcal{D}) = \prod_{k \in K} \mathcal{L}_k(\theta; \mathcal{D})^{w_k}$$

where $\mathcal{L}_k(\cdot)$ is the likelihood function resulting from the conditional probability (or density) function of a subset of

---

[2]Parameter $\theta_{ij}$ will not, of course, affect the likelihood function if $Y_i \leftrightarrow Y_j$ is not in $\mathcal{G}_m$. One could otherwise interpret $\theta_{ij}$ as simply not existing in this case, but this way of interpreting $\theta_{ij}$ will make the presentation easier without being less precise.

**Y** given another subset, and $w_k$ are user-specified weight parameters. It is intuitive to understand why a composite likelihood function can provide consistent estimates of $\theta$: if an unique model is identifiable from the marginal conditional densities used in each $\mathcal{L}_k(\cdot)$, then maximizing the log-composite likelihood is equivalent to minimizing the KL-divergence between each marginal conditional and the empirical distribution. By matching each marginal as the divergence goes to zero, one recovers the parameters.

Joreskog and Moustaki (2001) applied this concept in the context of latent variable models, by fitting a probit model using univariate and bivariate marginals with equal weight. A key fact is that bivariate likelihoods can be integrated numerically, since only two underlying latent variables are present. Although still computationally demanding, this approach does not require any Markov chain Monte Carlo within a stochastic EM procedure, nor requires biased approximations such as mean-field methods. We initially propose a similar idea, where our (penalized) composite likelihood function is given by

$$PCL(\mathcal{G}_m, \beta, \Sigma) \equiv \mathcal{F}_{\mathcal{G}_m}^{(\beta,\Sigma)} + \log \pi(\mathcal{G}_m), \quad (6)$$

$$\mathcal{F}_{\mathcal{G}_m}^{(\beta,\Sigma)} \equiv \sum_{i<j} \log \mathcal{P}(\mathbf{Y}_i^{1:N}, \mathbf{Y}_j^{1:N} \mid \mathcal{G}_m, \beta, \Sigma)$$

When using one-parameter copula functions (Nelsen, 2007), each bivariate term $\mathcal{P}(\mathbf{Y}_i^{1:N}, \mathbf{Y}_j^{1:N} \mid \mathcal{G}_m, \beta, \Sigma)$ requires the numerical integration of at most three terms: the copula parameter $\theta_{ij}$ corresponding to the edge $Y_i \leftrightarrow Y_j$ and up to two latent variables per data point configuration, as explained in Section 3.3.

A greedy procedure for optimizing $PCL(\cdot, \cdot, \cdot)$ is outlined in table Algorithm 1. It takes as input a partition over observed variables ($\mathcal{S}$) and a dataset $\mathcal{D}$. Algorithm 1 alternates between optimizing the objective function with respect to the continuous parameters (Step 5), and optimizing structure (Steps 6). Optimization in Step 5 is done with respect to the coefficient space $\Omega_\beta$ and correlation matrix space $\Omega_\Sigma$, as explained in further detail in Section 3.3. The structural update is a standard greedy algorithm that picks the best choice of graph within the graph space $\mathcal{G}_m^{+/-}$: the space that includes $\mathcal{G}_m$ and all mixed graphs that differ from $\mathcal{G}_m$ by exactly one bi-directed edge. Finally, procedure GETDAG($\mathcal{S}$) at the beginning of Algorithm 1 just returns the variable space ($\mathbf{Y}, \mathbf{X}_\mathcal{S}$) and the initial mixed graph without any bi-directed edges. The initialization of parameters is discussed in the Supplementary Material.

### 3.2 Learning via Distributed EM Bounds

Even for problems with large sample sizes, one might be concerned that using only pairwise regions might imply low statistical efficiency. Higher (statistical) efficiency can

---

**Algorithm 1** Pairwise Structured CCA Learning

1: **procedure** LEARNSTRUCTUREDCCA-I($\mathcal{S}, \mathcal{D}$)
2:     $\{\mathbf{Y}, \mathbf{X}_\mathcal{S}, \mathcal{G}_m\} \leftarrow$ GETDAG($\mathcal{S}$)
3:     $\{\beta, \Sigma\} \leftarrow$ INITPARAMETERS($\mathcal{G}_m, \mathcal{D}$)
4:     **repeat**
5:        $\{\beta, \Sigma\} \leftarrow \arg\max_{(\Omega_\beta, \Omega_\Sigma)} PCL(\mathcal{G}_m, \beta, \Sigma)$
6:        $\mathcal{G}_m \leftarrow \arg\max_{(\mathcal{G}_m^{+/-})} PCL(\mathcal{G}_m, \beta, \Sigma)$
7:     **until** $\mathcal{G}_m$ has not changed.
8:     **return** $\mathcal{G}_m$
9: **end procedure**

---

be obtained by using components with more than two observed variables. This is undesirable, as it might require sophisticated integration methods, including MCMC. We present a different way of sharing statistical power among different pairwise likelihood functions without requiring an integration procedure on dimensions higher than in the pairwise procedure of the previous Section. It is based on ideas adapted from the expectation-maximization (EM) family of optimization methods (Dempster et al., 1977).

Using the standard Jensen bound for convex combinations, but applied independently to different terms in a summation, it is possible to lower-bound $\mathcal{F}_{\mathcal{G}_m}^{(\beta,\Sigma)}$ as

$$\sum_{i<j} \int q_{ij}(\theta_{ij}) \log \frac{\mathcal{P}_{ij}(\mathbf{Y}_i^{1:N}, \mathbf{Y}_j^{1:N}, \theta_{ij} \mid \mathcal{G}_m, \beta, \Sigma)}{q_{ij}(\theta_{ij})} d\theta_{ij} \quad (7)$$

The bound holds for any choice of $q_{ij}(\cdot)$ and it is well-known to be maximized by choosing $q_{ij}(\cdot)$ to be $\mathcal{P}(\theta \mid \mathbf{Y}_i^{1:N}, \mathbf{Y}_j^{1:N}, \mathcal{G}_m, \beta, \Sigma)$ (Neal and Hinton, 1998).

We will further modify the idea in (7) with a different matching between functionals $q_{ij}(\cdot)$ and log-likelihood functions. Let $X_i \in \mathbf{X}_\mathcal{S}$ and recall $\mathcal{S}_i$ are the observed children of $X_i$ in **Y**. Let $|\mathcal{S}|$ be the number of latent variables. The trick is to first rewrite $\mathcal{F}_{\mathcal{G}_m}^{(\beta,\Sigma)}$ as in Equation (8) displayed in Table 1. Using an arbitrary set of distributions $\{q_{mn}(\cdot)\}, 1 \leq m < n \leq |\mathcal{S}|$, $PCL(\mathcal{G}_m, \beta, \Sigma)$ can then be rewritten and elementwise bounded as

$$PCL(\mathcal{G}_m, \beta, \Sigma) \geq \mathcal{Q}_{\mathcal{G}_m}^{(\beta, \Sigma, \{q_{mn}(\cdot)\})} + \log \pi(\mathcal{G}_m) + \kappa \quad (10)$$

where $\mathcal{Q}_{\mathcal{G}_m}^{(\beta,\Sigma)}$ is defined as Equation (9) in Table 1, and constant $\kappa$ does not depend on the free parameters $\{\mathcal{G}_m, \beta, \Sigma\}$.

Given this setup, we finally define $q_{mn}(\Theta_{mn})$ to be the joint distribution $\mathcal{P}(\Theta_{mn} \mid \mathbf{Y}_{mn}^{1:N}, \mathcal{G}_m, \beta, \Sigma)$, where $\mathbf{Y}_{mn}$ are the *joint children* of $X_m$ and $X_n$, and $\Theta_{mn}$ are the copula parameters used in the corresponding marginal model for $\mathbf{Y}_{mn}$. Function $q_{mn}(\theta_{ij})$ is then the corresponding (univariate) marginal of $q_{mn}(\Theta_{mn})$.

The desirable property of (10) is that, while it still requires only a three-dimensional integration (details in Section 3.3), information from $Y_k$, not in $\{Y_i, Y_j\}$, can be

Table 1: Components of a Pairwise Composite Likelihood Score Function

$$\mathcal{F}_{\mathcal{G}_m}^{(\beta,\Sigma)} = \sum_{m<n} \sum_{Y_i \in \mathcal{S}_m} \sum_{Y_j \in \mathcal{S}_n} \log \mathcal{P}(\mathbf{Y}_i^{1:N}, \mathbf{Y}_j^{1:N} \mid \mathcal{G}_m, \beta, \Sigma) + \sum_{m=1}^{|\mathcal{S}|} \frac{1}{|\mathcal{S}|-1} \sum_{n=1}^{|\mathcal{S}|-1} \sum_{\{Y_i,Y_j\} \subset \mathcal{S}_m} \log \mathcal{P}(\mathbf{Y}_i^{1:N}, \mathbf{Y}_j^{1:N} \mid \mathcal{G}_m, \beta, \Sigma) \quad (8)$$

$$\mathcal{Q}_{\mathcal{G}_m}^{(\beta,\Sigma,\{q_{mn}(\cdot)\})} = \sum_{m<n} \sum_{Y_i \in \mathcal{S}_m} \sum_{Y_j \in \mathcal{S}_n} \int q_{mn}(\theta_{ij}) \log \mathcal{P}(\mathbf{Y}_i^{1:N}, \mathbf{Y}_j^{1:N} \mid \mathcal{G}_m, \beta, \Sigma, \theta_{ij}) d\,\theta_{ij} + $$
$$\frac{1}{|\mathcal{S}|-1} \sum_{m=1}^{|\mathcal{S}|} \sum_{n \neq m} \sum_{\{Y_i,Y_j\} \subset \mathcal{S}_m} \int q_{mn}(\theta_{ij}) \log \mathcal{P}(\mathbf{Y}_i^{1:N}, \mathbf{Y}_j^{1:N} \mid \mathcal{G}_m, \beta, \Sigma, \theta_{ij}) d\,\theta_{ij} \quad (9)$$

---

**Algorithm 2** Modified Pairwise Structured CCA Learning
1: **procedure** LEARNSTRUCTUREDCCA-II($\mathcal{S},\mathcal{D}$)
2:    $\{\mathbf{Y}, \mathbf{X}_\mathcal{S}, \mathcal{G}_m\} \leftarrow$ GETDAG($\mathcal{S}$)
3:    $\{\beta, \Sigma\} \leftarrow$ INITPARAMETERS($\mathcal{G}_m, \mathcal{D}$)
4:    **repeat**
5:      **for** $1 \leq m < n \leq |\mathcal{S}|$ **do**
6:        $q_{mn}(\cdot) \leftarrow \mathcal{P}(\Theta_{mn} \mid \mathbf{Y}_{mn}^{1:N}, \beta, \Sigma, \mathcal{G}_m)$
7:      **end for**
8:      $\{\beta, \Sigma\} \leftarrow \arg\max_{(\Omega_\beta, \Omega_\Sigma)} \mathcal{Q}_{\mathcal{G}_m}^{(\beta,\Sigma,\{q_{mn}(\cdot)\})}$
9:      $\mathcal{G}_m \leftarrow \arg\max_{(\mathcal{G}_m^{+/-})} PCL(\mathcal{G}_m, \beta, \Sigma)$
10:    **until** $\mathcal{G}_m$ has not changed.
11:    **return** $\mathcal{G}_m$
12: **end procedure**

---

passed around. Consider the graph in Figure 1 again. There is a d-connecting path between $Y_2$ and $Y_6$ given $Y_4$ (or "m-connecting," using the nomenclature from Richardson and Spirtes, 2002). Hence, there is a three-way interaction between $Y_2, Y_4$ and $Y_6$ that varies according to the copula between $Y_2 \leftrightarrow Y_4$, and information from $\mathbf{Y}_6$ is propagated to the estimation of $\beta_{2;1}$ and $\beta_{4;1}$ via the distribution of $\theta_{2;4}$.

A modification of Algorithm 1 is shown in Algorithm 2. The parameter fitting procedure is now expanded into Steps 5-8, with the structure learning step unmodified. The posterior $\mathcal{P}(\Theta_{mn} \mid \mathbf{Y}_{mn}^{1:N}, \mathcal{G}_m, \beta, \Sigma)$ needs to be calculated efficiently and it has to lead to efficient integration in Step 8. This is discussed in Section 3.3. Moreover, it might be surprising that in Step 9 we do not optimize structure with respect to the same function of Step 8. Optimizing the bound of Equation (10) with respect to the graph is equivalent to a composite likelihood Structural EM procedure (Friedman, 1998). While Structural EM procedures are definitely useful, we prefer to not apply it uncritically to our special situation. EM is a coordinate ascent method (Neal and Hinton, 1998), and as such it is more prone to get stuck in a local maxima compared to optimizing marginal likelihoods directly. This is particularly true when optimizing with respect to a discrete structure – in this case, it is common to prematurely converge to a structure with fewer bi-directed edges than expected if we optimize the Structural EM lower bound starting with an empty bi-directed component. Since for each data point we have no more than two latent variables per log-likelihood term in $PCL(\cdot, \cdot, \cdot)$, we avoid maximizing a lower bound in Step 9. Convergence issues are discussed in the Supplementary Material.

### 3.3 Implementation Details

For Algorithm 2, we obtain the posterior distribution $q_{mn}(\Theta_{mn})$ using the Laplace approximation (MacKay, 2003). Our goal is to use Algorithms 1 and 2 in problems with large sample sizes, and hence such an approximation can provide a reasonably accurate replacement for the exact posterior – which has no closed form and would require a numerical method in any case.

Even the Laplace approximation procedure requires another inner approximation. Recall that in order to obtain the mean vector and covariance matrix required to approximate the distribution of $\Theta_{mn}$ with a Gaussian, we need to maximize $\log \mathcal{P}(\mathbf{Y}_{mn}^{1:N} \mid \mathcal{G}_m, \beta, \Sigma, \Theta_{ij}) + \log \pi(\Theta_{mn})$. Let $P$ be the total number of children between $X_i$ and $X_j$. Then $\mathcal{P}(\mathbf{Y}_{mn}^{1:N} \mid \Theta_{mn}, \mathcal{G}_m, \beta, \Sigma)$ can be rewritten as

$$\prod_{\mathbf{y}_{mn} \in \{0,1\}^P} \left( \int \mathcal{P}(\mathbf{y}_{mn}, x_m, x_n \mid \Theta_{mn}, \dots) d\,x_m d\,x_n \right)^{N_{mn}}$$

which takes constant time in sample size[3] given pre-computed sufficient statistics $N_{mn}$, the count of events $\mathbf{Y}_{mn} = \mathbf{y}_{mn}$ in dataset $\mathcal{D}$, and under an approximation to each of the two-dimensional integrals.

In our implementation, we use an approximation for the

---
[3]Assuming that the marginal of $\mathcal{G}_m$ over $\mathbf{Y}_{mn}$ has a bounded tree-width, this expression is tractable in the dimensionality of the problem using the CDN inference method to obtain the likelihood function (Huang and Frey, 2011, see also the Supplementary Material). The full graph $\mathcal{G}_m$ might have a large tree-width, as long as the subgraph given by $\mathbf{Y}_{mn}$ does not.

integral of $\mathcal{P}(\mathbf{y}_{mn}, x_m, x_n \mid \Theta_{mn}, \beta, \Sigma, \mathcal{G}_m)$ of the form

$$\approx \sum_k w_k(\sigma_{mn}) \mathcal{P}(\mathbf{y}_{mn}, x_m^{(k)}, x_n^{(k)} \mid \Theta_{mn}, \beta, \Sigma, \mathcal{G}_m) \tag{11}$$

for a fixed set of grid points $\{(x_m^{(k)}, x_n^{(k)})\}$ regardless of the values of $m$ and $n$, but a function of the covariance $\sigma_{mn} \equiv (\Sigma)_{mn}$. Any quadrature method could be used.

We opted for an admittedly crude but simple implementation which uses an arguably excessive number of points compared to adaptive quadrature methods, but which is relatively amortized taken into account such integrations need to be done several thousand times – we need to calculate our weights $w_k$ only once. Without loss of generality, $\Sigma$ can be assumed to be a correlation matrix. For each marginal $\{x_m, x_n\}$, we naively space the grid points uniformly between the 0.01 and 0.99 quantiles of the standard Gaussian. The result is a division of the space into squares centered at each $\{(x_m^{(k)}, x_n^{(k)})\}$ which does not depend on $m$ and $n$. We define $w_k(\sigma_{mn})$ to be the probability mass of each square. This can be pre-computed only once in the whole procedure, also using the following simplification of $\Sigma$: we allow each entry $\sigma_{mn}$ to lie only within the discrete set of choices $\{-0.99, -0.97, \ldots, 0.97, 0.99\}$ and cache $w_k(\sigma_{mn})$ for every $k$ and possible value of $\sigma_{mn}$. Since they correspond to bivariate Gaussian integrals, recomputing these weights $\mathcal{O}(p^2)$ times at each iteration would otherwise be very costly.

Equations (8) and (9) require integrals over $\theta_{mn}$. Given the approximation for the bivariate likelihood of each pairwise data point configuration, we introduced a discrete approximation for the prior $\pi(\theta_{mn})$ (Algorithm 1) or univariate Gaussian marginal $q_{mn}(\theta_{mn})$ that results from the Laplace approximation (Algorithm 2). The grid points are naively a number of quantiles corresponding to uniformly spread cumulative probabilities in $[0, 1]$. These quantiles are recomputed at every iteration, since this is relatively cheap.

When optimizing parameters, we optimize $\beta$ for a fixed $\Sigma$ by gradient methods[4]. We then optimize $\Sigma$ given $\{\beta_{i1}, \beta_{i0}\}$ without enforcing positive definiteness: the evaluation of $\mathcal{F}$ and $\mathcal{Q}$ does not require this condition. The objective functions decouple over the entries of $\Sigma$ and hence each entry $\sigma_{mn}$ can be optimized separately – by searching over the discretized space $\{-0.99, -0.97, \ldots, 0.97, 0.99\}$, as required to allow for the caching of $w_k(\sigma_{mn})$.

### 3.4 Bivariate Residual Search

An alternative to the expensive iterative methods from the previous section is to do a single hypothesis test for each

---

[4]In order to reduce the number of parameters, we calculate each intercept $\beta_{i0}$ as a function of the slope $\beta_{i1}$ such that the marginal probability $\mathcal{P}(Y_i = 0 \mid \beta, \Sigma)$ matches the empirical probability. Hence the only free parameters are the slopes $\{\beta_{i1}\}$.

---

**Algorithm 3** Single-shot Structured CCA Learning
1: **procedure** LEARNSTRUCTUREDCCA-0($\mathcal{S}, \mathcal{D}$)
2:    $\{\mathbf{Y}, \mathbf{X}_\mathcal{S}, \mathcal{G}_m\} \leftarrow$ GETDAG($\mathcal{S}$)
3:    $\{\beta, \Sigma\} \leftarrow \arg\max_{(\Omega_\beta, \Omega_\Sigma)} PCL(\mathcal{G}_m, \beta, \Sigma)$
4:    **for** $1 \leq i < j \leq |\mathbf{Y}|$ **do**
5:       $\mathcal{G}_m^{ij} \leftarrow \arg\max_{(\mathcal{G}_m^{ij(+/-)})} PCL(\mathcal{G}_m, \beta, \Sigma)$
6:    **end for**
7:    **return** $\cup_{ij} \mathcal{G}_m^{ij}$
8: **end procedure**

bi-directed edge. If one knew parameters $\{\beta, \Sigma\}$, a possibility is to do a $\chi^2$-like measure of fitness of the implied bivariate contingency table for each $\{Y_i, Y_j\}$, and add the corresponding bi-directed edge if there is evidence of misfit. Without knowledge of $\{\beta, \Sigma\}$, a possibility is to fit the model without bi-directed edges as a surrogate, in the hope that such estimates are good enough to detect misfit. In Algorithm 3, we outline a procedure where the test of misfit is to choose between two models with the single edge $Y_i \leftrightarrow Y_j$ and none (our definition of the space $\mathcal{G}_m^{ij(+/-)}$) based on $PCL(\cdot, \cdot, \cdot)$ – essentially a Bayesian test for the edge in the bivariate model $\mathcal{P}(Y_i, Y_j \mid \beta, \Sigma)$. The final graph is given by the union of all edges that were selected by the bivariate tests. The shortcoming of course is that the initial parameter estimate might be bad if there are reasonably strong dependencies due to the unaccounted latent variables. Iterating the procedure, however, gives a procedure that is essentially Algorithm 1, albeit with a "parallel" testing of the edges. For simplicity, we do not consider the continuum between sequential and parallel tests (e.g., methods where initial iterations would modify one edge only per parameter update, allowing multiple edge modifications later on) and evaluate Algorithm 3 only.

## 4 RELATED WORK

More conventional uses of EM in the context of composite likelihod have been discussed by Varin et al. (2011) and Gao and Song (2011). CCA has been applied in other contexts in machine learning, such as analysing text data under different languages (Hardoon et al., 2004). This setup and usage is very different from our motivation, which focus on applications with several small sets of measurements, and where the special structure of the latent space (a one-to-one association between latent variables and groups of observable variables) is motivated by particular applications in measurement error problems (Bollen, 1989). Concerning model selection in structured latent spaces, Silva and Ghahramani (2009) provides an approach based on Gaussian distributions, and a Gaussian copula method can be readily applied to the binary modeling case. However, we want to avoid a full likelihood approach for scalability reasons. While Silva and Ghahramani (2009) provides priors over sparse covariance matrices, it is not clear what the marginals of such priors are over subsets of the observations: sparse inverse Wishart priors do not have a closed form for their marginals. This makes the Bayesian Gaussian approach seemingly hard to apply in the composite likelihood scenario. The use of composite likelihood methods in model selection (Varin and Vidoni, 2005) has been increasing in recent years, including variations where parameters across different likelihood components do not need to be constrained to be the same: a postprocessing method can be used to combine estimates from different regions into a single point estimate (e.g., Meinshausen and Buehlmann, 2006). Further consequences of ignoring association due to unspecified factors in a latent variable model are discussed by Westfall et al. (2012). It is also important to mention that the method here introduced generalizes the learning methods of Huang and Frey (2011), since the CDN is a special case. It is also an alternative to maximum likelihood estimation for networks of large tree-width.

## 5 EVALUATION

We evaluate the approach with a synthetic and a real-world experiment. The prior over graphical structures is given by an independent probability of 0.1 for each bi-directed edge. We use Frank copulas for the dependence structure (Nelsen, 2007). The prior for parameter $\theta_{ij}$ is defined by first sampling a standard Gaussian $Z \sim \mathcal{N}(0,1)$ and squashing it as $\theta_{ij} = [1/(1+e^{-z})] \times 50 - 25$, which generates $\theta_{ij}$ in the interval $[-25, 25]$ – beyond which the Frank copula gives numerically unstable results in our code.

### 5.1 Synthetic Studies

We simulate[5] 20 networks with 4 latent variables and 4 children per latent variable. Bi-directed edges were added independently with probability 0.2. The network is then pruned randomly so that each latent variable d-separates at least three children, and that each observed variable has no more than 3 adjacent bi-directed edges. The average number of resulting edges was approximately 18. Results for three evaluation measures are shown in Figure 2. We compare the three methods of Section 3, doing a comparison at three different sample sizes (1000, 5000 and 10000). We show the average paired difference over 20 trials between

---
[5]Other details about the simulation: a "signal factor" (the ratio between the variance $\beta_{i1}^2$ implied by the Gaussian $X_m$, and $\beta_{i1}^2 + 1$ given by the added variance of the probit model) is the sampled uniformly within the interval $[0.2, 0.6]$ for each variable $Y_i$. Slope $\beta_{i1}$ is then set accordingly as a function of the signal factor, with its sign chosen with probability 0.5. A "marginal factor" (the marginal probability $\mathcal{P}(Y_i = 0)$) is uniformly sampled within the interval $[0.1, 0.9]$, and intercept $\beta_{i0}$ is set according to the marginal factor and the sampled value of $\beta_{i1}$. We sample each copula parameter independently, uniformly in the interval $[10, 15]$. Finally, we generate $\Sigma$ by rescaling as a correlation matrix the result of $\sum_{k=1}^{4} z_k z_k^{\mathsf{T}}$, where $Z_K \sim \mathcal{N}(0,1)$.

LEARNSTRUCTUREDCCA-II (LSC-II) and LSC-I, and between LSC-II and LSC-0. The first criterion is the root mean squared error of the average slope coefficients $\{\beta_1\}$ with respect to the true model; the second criterion is "edge omission," the number of incorrectly removed edges divided by the total number of edges; the third criterion is "edge commission," incorrect addition of an edge divided by the total number of possible additions. The number of times where the difference is positive with the corresponding p-values for a Wilcoxon signed rank test are given below (stars indicate numbers less than 0.05):

|           | 1000 |      | 5000 |      | 10000 |      |
|-----------|------|------|------|------|-------|------|
| **Slope**     | I    | 0    | I    | 0    | I     | 0    |
| number    | 13   | 6    | 17   | 15   | 15    | 13   |
| p-value   | 0.22 | 0.25 | *    | *    | *     | 0.06 |
| **Omission**  | I    | 0    | I    | 0    | I     | 0    |
| number    | 11   | 18   | 6    | 14   | 6     | 9    |
| p-value   | 0.17 | *    | 0.82 | *    | 0.62  | 0.22 |
| **Commision** | I    | 0    | I    | 0    | I     | 0    |
| number    | 5    | 2    | 15   | 16   | 16    | 18   |
| p-value   | 0.28 | *    | *    | *    | *     | *    |

The upshot is that Algorithm 2 does at least as well as the others concerning edge omission, but with substantially fewer false positives – a problem that particularly affects the single-shot algorithm. Algorithm 2 is also more robust concerning parameter estimation. It has to be said, however, that all methods do badly concerning edge omission at sample size 1000: in more than 10 trials, Algorithm 2 had edge omission rates over 0.5. At sample size 5000, this substantially decreased (17 trials under 0.2, 10 under 0.15). Algorithm 2 commission errors are typically low ($< 0.05$) for all datasets. It is also relevant that Algorithm 2 typically converges faster than Algorithm 1 despite the extra step on approximating posteriors with the Laplace approximation.

### 5.2 Analysis of the NHS Staff Survey

We present a simple application to the modeling of response patterns in the 2009 National NHS Staff Survey (Care Quality Comission and Aston University, 2010). We now emphasize less an evaluation of the structure and more the initial goal described in Section 2.2 on showing how finding bi-directed structures helps in estimating functionals of $\mathcal{P}(\mathbf{X}_\mathcal{S} \mid \mathbf{Y})$. The NHS (National Health Service) is the public healthcare system of the United Kingdom. A survey concerning different aspects of job satisfaction, work conditions, training and other factors takes place regularly. We provide an analysis of the 2009 survey, which was returned by 156,951 respondents. Under the assumption that sections in the questionnaire translate to particular trait factors, we evaluate how much is gained by allowing a latent variable model to be adaptive to external factors not originally included in the model. Responses to questions are either binary Yes/No questions or encoded in an ordinal scale (varying in 5 points from strong disagreement to strong agreement). We en-coded ordinal data as binary, assigning the value of 1 to "Agree" and "Strongly agree" responses, and 0 otherwise (including missing or not applicable responses). We generated 9 factors out of the questionnaire with a total of 50 observed variables. The selection and partitioning process is described in detail in the Supplementary Material. We randomly selected 100,000 respondents as a training set and fit a model using LEARNSTRUCTUREDCCA-II. A total of about 40 bi-directed edges was generated. We also fit the model without any bi-directed edges. Since we cannot calculate marginal likelihoods easily as a way of comparing the models, we resort to evaluating the predictive ability of the latent representations. Given these two models, we generate latent embeddings of the observations[6] in the test set by maximizing $\sum_{ij} \log \mathcal{P}(Y_i^{(d)}, Y_j^{(d)}, X_i^{(d)}, X_j^{(d)} \mid \beta, \Sigma, \mathcal{G}_m)$, for each data point $d$, with respect to the latent variables. Here $X_i$ and $X_j$ are the (possibly unique) latent parents of $Y_i$ and $Y_j$. This results in over 50,000 points in a 9-dimensional space. Half of these points were used to build 11 logistic regression models to predict answers to questions not included in the model[7]. We calculated the area under the curve for predictions done in the test set of approximately 28,000 points. Results are shown in the Table 2, which illustrates that the mixed model does at least as well or better at generating a latent representation of the observations, while preserving the interpretability of the model.

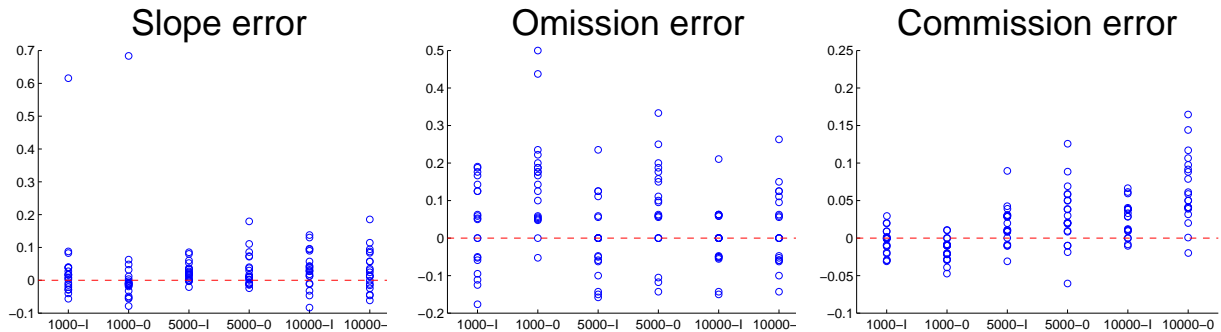

Figure 2: Differences between LEARNSTRUCTUREDCCA-II (LSC-II) and LSC-I are labeled, at three different samples sizes, as 1000-I, 5000-I and 10000-I. Differences between LSC-II and LSC-0 are labeled analogously.

## 6 CONCLUSION

We summarize the main features of our method:

- it searchs only over a projection of an infinite dimensional latent space into a mixed graph structure,

Table 2: Comparison of the mixed graph CCA model (MCCA) against the standard, fixed structure, model (SCCA) in 11 binary classification tasks, as measured by the area under the curve.

|     | MCCA | SCCA |     | MCCA | SCCA |
|-----|------|------|-----|------|------|
| Q1  | 0.71 | 0.71 | Q7  | 0.80 | 0.79 |
| Q2  | 0.75 | 0.75 | Q8  | 0.82 | 0.81 |
| Q3  | 0.86 | 0.82 | Q9  | 0.86 | 0.83 |
| Q4  | 0.90 | 0.82 | Q10 | 0.69 | 0.69 |
| Q5  | 0.79 | 0.80 | Q11 | 0.78 | 0.75 |
| Q6  | 0.73 | 0.72 |     |      |      |

instead of explicitly adding latent variables. Conditioned on $\mathbf{X}_\mathcal{S}$, the composite likelihood function can be calculated analytically. This is possible due to the *implicit latent variable representation* of the CDN;
- assuming no missing data, it requires only the sufficient statistics for regions of bounded size, computable with a single pass through the data, possible due to the *marginalization property* of the CDN;
- optimization is unconstrained, thanks to the *parameter modularity* property of the copula formulation;
- it allows for the use of simple deterministic integration methods, while still providing a mechanism to propagate information beyond simple pairs of observed variables – a *novel variation of composite likelihood estimation*, to the best of our knowledge.

Although the approach scales well in the sample size, the parameter fitting step is still particularly expensive in high-dimensions, as also noted by Joreskog and Moustaki (2001). A more efficient algorithm can be achieved by a smarter implementation of the gradient optimization method, or by relaxing the restriction that parameter estimates need to be the same across different likelihood components (Meinshausen and Buehlmann, 2006). Moreover, having an adaptive structure for the partition of the random variables is also an important direction.

---
[6] After fitting both models, 5 of the observed variables were given extreme coefficients in their measurement equations (in both models). In order to make the embedding process numerically stable, we removed these 5 variables from the testing.

[7] Questions concerned job satisfaction. See Supp. Material.

**Acknowledgements**

The author would like to thank the three anonymous reviewers for very many helpful comments that inspired improvements in the presentation of the method and results.